# THE ROLE OF ARTIFICIAL INTELLIGENCE TECHNOLOGIES IN CRISIS RESPONSE

Khaled M. Khalil, M. Abdel-Aziz, Taymour T. Nazmy & Abdel-Badeeh M. Salem

Faculty of Computer and Information Science, Ain shams University Cairo, Egypt
kmkmohamed@gmail.com, mhaziz@aucegypt.edu, ntaymoor@yahoo.com, absalem@asunet.shams.edu.eg

*Abstract:* Crisis response poses many of the most difficult information technology in crisis management. It requires information and communication-intensive efforts, utilized for reducing uncertainty, calculating and comparing costs and benefits, and managing resources in a fashion beyond those regularly available to handle routine problems. In this paper, we explore the benefits of artificial intelligence technologies in crisis response. This paper discusses the role of artificial intelligence technologies; namely, robotics, ontology and semantic web, and multi-agent systems in crisis response.

## 1.0 Introduction

Crisis events, like the 9.11 attack, Hurricane Katrina and the tsunami devastation, have dramatic impact on human society, economy and environment. The crisis response term is defined as the immediate protection of property and life during the crises events to reduce deaths and injuries. Crisis response requires urgent action and the coordinated application of resources, facilities, and efforts. It includes actions taken before the actual crisis event (e.g., hurricane warning is received), in response to the immediate impact of a crisis, and as sustained effort during the course of the crisis. Depending upon the magnitude and complexity of the crisis, response may be a large-scale and multi-organizational operation involving many layers of authorities, commercial entities, volunteer organizations, media organizations, and the public. These entities work together as a virtual organization to save lives, preserve infrastructure and community resources, and reestablish normalcy within the community [1]. Artificial intelligence technology tries to improve the efficiency of the management process during the crisis response via: robotics sustaining urban search and rescue operations [12], enhancing information sharing using ontologies [5], providing customized query to crisis actors [3], and providing multi-agent systems for real time support [15] and simulated environments [8]. We will discuss these technologies and those roles in crisis response.

First, the diversity structure of crisis area, rescuers safety and the necessity of quickly and reliably examining targeted regions forces rescue agencies to use multi-robot solutions in the field of urban search and rescue. Robots provide variety of functions in the crisis context, such as area exploration, mapping and expediting the search for victims. One of the first uses was "VGTV and MicroTracs" robots, which are used during the World Trade Center crisis in New York [12] to search for victims under collapsed buildings. Successively, aerial robots ("T-Rex helicopter" from Like90) are used at Hurricane Katrina and boat robots ("AEOS-1") are used at Hurricane Wilma.

Second, from the point of view of information processing, the success of crisis response largely depends on gathering information from distributed sources, integrating it and then making decisions. It is clear that such complexity makes it impossible for any single human or even a team to fulfill the roles adequately [3]. Ontologies and semantic web are adopted to solve integrating problems, for example ontologies are used in integrating heterogeneous information sources and semantic web services are used to provide customized queries to crisis actors. The World Wide Web Consortium (W3C) [4] and E-response project represent the noticeable effort in the way of building crisis response ontologies and getting the benefits of semantic web services. W3C focused on identifying and building standard ontology for crisis response, while E-response project focused on building overall crisis response ontology and semantic web services based on the created ontologies.

Third, crisis response problems are not solvable by single responder and a heterogeneous team is needed. Heterogeneous team needs planning and coordination capabilities to complete his mission successfully. A multi-agent system provides the decisive solution to all problems related to interaction and coordination of response teams. Related multi-agent systems for crisis response include real-time support and simulation systems such as DrillSim [8], DEFACTO [15] and WIPER [14].

In the following sections we discuss in details artificial intelligence technologies: robots, ontologies and semantic web, and multi-agent systems contributions in crisis response.

## 2.0 Robotics

Robotics is a growing research area in crisis response. Multi-robot solutions had been adopted in a wide range of crisis response operations. Specifically, robots are used in Urban Search and Rescue (USAR) operations. Urban Search and Rescue involves locating, rescuing, and medically stabilizing victims trapped in confined spaces. USAR workers have 48 hours to find trapped survivors in a collapsed structure; otherwise the likelihood of finding victims still alive is nearly zero. Greer [7] had summarized challenges that USAR team have to overcome into four areas, (1) efficient response, (2) rescuers safety, (3) environment disturbance and climatic conditions, and (4) inappropriate equipment and resources. Buildings debris prevents rescue workers from searching due to the unacceptable personal risk from further collapse, besides collapse confined create spaces which are frequently too small for people to enter limiting the search to no more than a few feet from the exterior. Rescuers may be crushed by structural collapse or may be suffered respiratory injuries due to hazardous materials, fumes and dust. The site needs to be shored up and made safe for rescuers to enter which takes up three to four critical early hours of the crisis which are crucial for finding victims alive.

Robots can bypass the danger and expedite the search for victims immediately after a collapse. Their ability to navigate through tightly confined spaces which people cannot access makes them extremely useful for quickly getting to a location within the crisis site. Robots can be deployed to a large crisis to search multiple locations simultaneously to expedite the search process. They can map the area and identify the location of victims using Radio Frequency Identification (RFID) tags. During the search they can deposit radio transmitters to be able to communicate with victims, use small probes to check victim's heart rate and body temperature and supply heat source and small amounts of food and medication to sustain the survivors [12]. One of the first uses of robots in search and rescue operation was during the World Trade Center crisis in New York. Figure 1 shows VGTV and MicroTracs by Inuktun robot used in rescue operations during the World Trade Center crisis in New York [12]. Micro-VGTV or Variable Geometry Tracked Vehicle can alter its shape during operation. The tracks, in their lowered configuration, take the shape of conventional crawler tracks. When the geometry is varied to the point where the vehicle is in its raised configuration, the tracks take the shape of a triangle. This unique feature allows the vehicle to negotiate obstacles, and operate in confined spaces and over rough terrain.

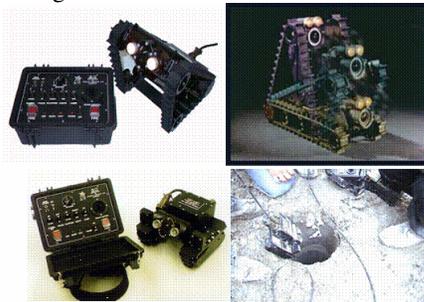
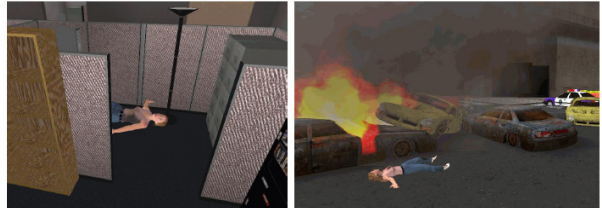

Figure 1: The Micro VGTV System by Inuktun with its control units

Figure 2: Representative snapshot of USARSim MicroTracs

Urban rescue and search simulation (USARSim) plays another vital role in crisis response. USARSim is a benchmark for evaluating robot platforms for their usability in crisis response. USARSim framework provides a development, testing and competition environment that is based on a realistic depiction of conditions after a real crisis, such as an earthquake or a major fire. Robots are simulated on the sensor and actuator level based on social behavior, making a transparent migration of code between real robots and their simulated counterparts possible, Figure 2 for example, a real robot may be exploring environment in cooperation with a virtual robot. The robots share map information and even see each other in their own respective representations of the real or virtual worlds.

## 3.0 Ontology and Semantic Web

Information management and processing in crisis response aimed to produce digital representations for a common response operational picture. This common picture cannot be effective without overcoming the following challenges [9]: (1) Diversity of information sources: information relevant to decision making may be dispersed from sensors where data is generated, to heterogeneous databases belonging to autonomous organizations. In addition, critical information may span various modalities, e.g., voice conversations among crisis responders, cameras data, sensor data streams, GIS(Geographical Information Systems)-oriented data and relational information in databases, (2) Diversity of information users: different people/organizations have different needs and urgency levels regarding the same information. According to theses challenges different sorts of data are used, but a common core set may be shared throughout. This common core set of information can be represented by ontology.

According to W3C [4] definition of crisis response ontology, crisis response ontology must describe the following critical steps:
- Once crisis is widely anticipated, sharing of data describing response and resource characteristics are needed.

- As the crisis unfolds gathering of data on its scope and emerging effects.
- As the response begins, gathering of data on its outages and missing links and matching with relief capacity.
- As the response by first responders is overwhelmed, sharing relief requests to prioritize relieving the first responders who are most overloaded or tired.
- As the relief unfolds gathering and integrating data from all responders to build a common baseline map of the situation and facilitate probes and first attempts at proactive data gathering.
- Characterizing problem states as chaotic (no baseline and no reliable map), complex (changing too fast to identify causes, requires probes) or manageable.
- Rapidly deploying compatible information and communication systems to local authorities and institutions capable of dealing with the manageable situations.
- Calling for expert review of action proposals to limit/contain chaotic situations, and mass peer review of probes that better define complex ones, with intent to limit the unanticipated side effects of management decisions.
- Comparing predicted to measured effects of interventions within 48-72 hours.
- Identifying situations which are not improving and calling for more options or more resources.
- Helping experienced response teams move on to the more complex situation by facilitating rapid handoff and just-in-time training of those less experienced.
- Guiding recovery and reconstruction efforts by identifying those outages or problems that most inhibit the resilience networks and outside relief efforts.
- Guiding resilience efforts by identifying which prevention and anticipation options (e.g. evacuation) could have prevented the most morbidity or loss of life-sustaining infrastructure.
- Passing off all data gathered in the disaster to the appropriate authority after the crisis passes, updating databases of vulnerable persons and places.

Different types of ontologies have been developed such as:
1. Ontologies for overall crisis response: E-response project has developed different types of response ontologies, such as ontology for overall crisis response process, pathology ontology, and healthcare ontology.
2. Robot ontology for urban search and rescue: Schlenoff [13] has developed robot ontology to capture relevant information about robots and their capabilities to assist in the development and testing of effective technologies for sensing, navigation, planning, integration, and human operator interaction within search and rescue robot systems. Captured information recognized in three categories: structural characteristics (such as size, weight, power source, locomotion mechanism, sensors and processors), functional capabilities (such as weather resistance, degree of autonomy, capabilities of locomotion, sensors and operations, and communications), and operational considerations (such as human operator training and education).
3. Decision making ontologies: Bloodsworth [3] has described COSMOA an ontology-centric multi-agent system that is aimed at supporting hospitals during the response to a large-scale incident event by producing a web-based emergency plan. COSMOA has been designed to support the decision-making process during the medical response. It is based on ontology layer which is simply a collection of one or more domain specific and generic ontologies. Ontologies are used within COSMOA to collect, integrate, reason on heterogeneous data (potential number of casualties and likely injuries) and then generate response plans. Response plans are posted on the response website, which are reviewed by crisis manager, responders and decision makers.

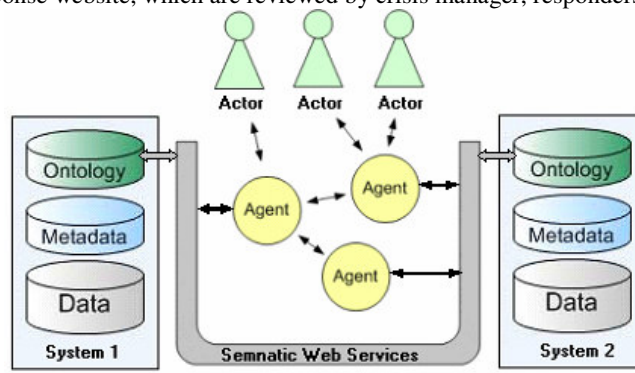

Figure 3: Ontologies deployed on legacy systems and semantic web services

Based on the deployed ontology, semantic web services are prototyped that provide data to crisis actors, Figure 3. Emergency Management Application (EMA) is an example of developed semantic web services [16]. EMA system has been designed to enable data and functionalities provided by existing legacy systems to be exposed as Web Services (WS). This system involves number of ontologies required to gather information from different sources. Based on those embedded ontologies, emergency officer can retrieve, process, display, and interact with only emergency relevant information more quickly and accurately.

The Semantic web vision of a crisis system that could answer a complicated request at the time of a crisis is far from realized. For example, if an emergency officer needed enough tents and food for 3400 people, deliverable in one day, first by air to the local city, then by road to the crisis area accompanied by fifteen distribution experts, the parts of this request would need at present to be broken into separate items. The required number of tents and amount of food would have to be computed, the location of the items discovered, and the logistics put in place. This would be done by building an ontology allowing machine inference in this domain [6].

## 4.0 Multi-Agent Systems

A multi-agent system (MAS) is a system composed of multiple interacting intelligent software agents. Multi-agent systems can be used to solve problems which are difficult or impossible for an individual agent to solve such as crisis response, and modeling social structures. Currently, multi-agent architecture is the essence of response systems. The original idea comes out from agent characteristics in MAS, such as autonomy, local view of environment, and capability of learning, planning, coordination and decentralized decision making. If we imagine that an agent can represent a crisis responder, so we can build a crisis response system based on agents' interaction and coordination. Agents can help crisis responders doing their planning, and coordination tasks or even replacing human in information gathering and specific decision making tasks. Another important research field in crisis response is the agent-based modeling and simulation, which are currently used for responders training and systems testing. DrillSim [8], DEFACTO [15] and WIPER [14] are examples of multi-agent systems for crisis response. We will discuss each system in brief, and table 1 includes comparison of the three systems:

(1) DrillSim is an augmented reality multi-agent simulation environment for testing IT solutions. The purpose of this environment is to play out a crisis response activity where agents might be either computer agents or real people playing diverse roles. An activity in DrillSim occurs in a hybrid world that is composed of (a) the simulated world generated by a multi-agent simulator and (b) a real world captured by a smart space. In order to capture real actors in the virtual space, DrillSim utilizes a sensing infrastructure that monitors and extracts information from real actors that is needed by simulator (such as agent location, agent state, etc.); in other words, DrillSim infuses actions and state of human actors in the virtual space. DrillSim modeled agent behavior (Figure 4) as a discrete process where agents alternate between sleep and awake states. Agents wake up and take some action every t time units. For this purpose, an agent acquires awareness of the world around it (i.e. event coding), transforms the acquired data into information, and makes decisions based on this information. Then, based on the decisions, it (re)generates a set of action plans. These plans dictate the actions the agent attempts before going to sleep again. For example, hearing a fire alarm results in the decision of exiting a floor, which results in a navigation plan to attempt to go from the current location to an exit location and force the agent trying to walk one step following the navigation plan.

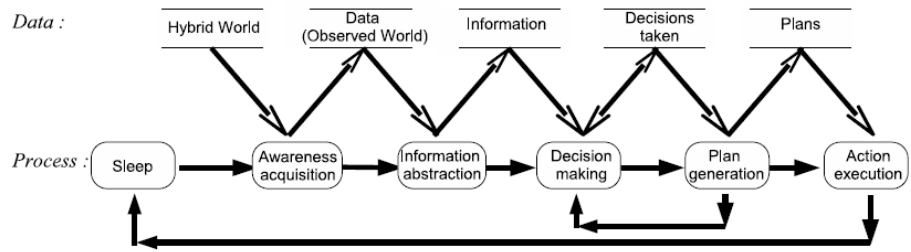

Figure 4: DrillSim Agent Behavior Process [8]

(2) DEFACTO (Demonstrating Effective Flexible Agent Coordination of Teams through Omnipresence) incorporates state of the art artificial intelligence, 3D visualization and human-interaction reasoning into a unique high fidelity system for training responders. By providing the responders interaction with the coordinating agent team in a complex environment, the responder can gain experience and draw valuable lessons that will be applicable in the real world. The DEFACTO system achieves this via (Figure 5): (i) omnipresent viewer – intuitive interface, (ii) and flexible interaction between the responder and the team. First, the 3D visualization interface enables human virtual omnipresence in the environment, improving human situational awareness and ability to assist agents. Second, generalizing past work on adjustable autonomy, the DEFACTO agent team chooses among a variety of "team-level" interaction strategies, even excluding humans from the loop in extreme circumstances. DEFACTO is comprised of various transfer-of-control strategies. Each transfer-of-control strategy is a preplanned sequence of actions to transfer control over a decision among multiple entities, for example, an $A_TH_1H_2$ strategy implies that a team of agents ($A_T$) attempts a decision and if it fails in the decision then the control over the decision is passed to a human $H_1$, and then if $H_1$ cannot reach a decision, then the control is passed to $H_2$.

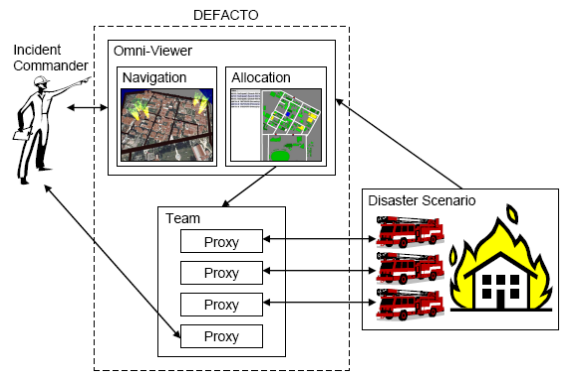

Figure 5: DEFACTO system applied [15]

(3) Wireless Phone-based Emergency Response (WIPER) system is designed to provide emergency planners and responders with an integrated system that will help to detect possible incident events, as well as to suggest and evaluate possible courses of response action. The system is designed as a distributed multi-agent system using web services and the service oriented architecture. WIPER is designed to evaluate potential plans of action using a series of GIS-enabled agent-based simulations that are grounded on real-time data from cell phone network providers. The system will interface with the existing cellular telephone network to allow cell phone activity to be monitored in aggregate, essentially creating a large scale, ad-hoc sensor network. The stream of incoming data will be monitored by an anomaly detection algorithm; flagging potential crisis events for further automated investigation. WIPER Agent-based simulations will attempt to predict the course of events and suggest potential mitigation plans, while displaying output at every level to human planners so that they can monitor the current situation, oversee the software process and make decisions.

WIPER architecture is composed of three layers: (1) Data Source and Measurement, (2) Detection, Simulation and Prediction, and (3) Decision Support. The Data Source and Measurement layer handles the acquisition of real time cell phone data, as well as the fixed transformations on the data, such as the calculation of triangulation information for providing more accurate location information on legacy handsets. The Detection, Simulation and Prediction layer analyzes incoming data for anomalies, attempts to simulate the anomaly to predict possible outcomes and suggests actions to mitigate the event. The Simulation and Prediction System will initially be used to predict simple movement and traffic patterns. Finally, the Decision Support layer presents the information from the other layers to end users in terms of summaries of traffic information for commuters, real time maps and simulations on the anomaly to first responders and potential plans for crisis planners. Table 1, compares among the three systems based on their objectives, architecture, application domain and features.

Table 1: Features of the three systems DrillSim, DEFACTO, and WIPER

| Criteria | Objectives | Architecture | Application Domain | Features |
|---|---|---|---|---|
| **DrillSim** | -Test-bed for IT Solutions | -Multi-Agent simulation | -(Fire) Floor Evacuation Simulation | -Micro level - every agent simulates and interacts with a real person.<br>-Agent Learning using recurrent artificial neural networks. |
| **DEFACTO** | -Address limitations of: (i) human situational awareness and (ii) the agent team's rigid interaction strategies. | -Software proxy architecture (Machinetta) and 3D visualization system | -Fire RoboCup rescue | -Omnipresent Viewer.<br>-Proxy Framework.<br>-Flexible Interaction.<br>-Adjustable Autonomy. |
| **WIPER** | -Evaluate potential plans of action using a series of GIS enabled Agent-Based simulations | -Web Service and Service Oriented Architecture + Multi-Agent System Design | -Building large scale ad-hoc sensor network based on the existing cellular telephone network.<br>-The Simulation and Prediction System will initially be used to predict simple movement and traffic patterns. | -Data stream will be monitored by an anomaly detection algorithm flagging potential crisis events.<br>-Agent-Based simulations will attempt to predict the course of events and suggest potential mitigation plans. |

## 5.0 Conclusion

Artificial intelligence techniques offer potentially powerful tools for the development of crisis response and management systems. The technologies of robotics, ontology and semantic web, and multi-agent systems can be useful to solve the problems of crisis response. This paper discusses the role of artificial intelligence technologies in crisis response. Robotics can be useful in urban search and rescue to bypass challenges faced by rescue workers and to expedite the search operations. Ontologies concepts and semantic web offers as many advantages in solving systems integration and interoperability problems. Ontologies are also used to collect, integrate and reason on heterogeneous information sources. Multi-agent systems features and methodologies is the core of crisis response systems. Crisis response systems take advantage of coordination and planning capabilities of multi-agent systems to handle response teams' coordination and interaction problems.